%% file: eccv2022submission.tex
\documentclass[runningheads]{llncs}
\usepackage{graphicx}
\usepackage{pifont}%

\newcommand{\titleLong}{ShAPO: Implicit Representations for Multi-Object Shape, Appearance, and Pose Optimization}%

\newcommand{\titleShort}{ShAPO }%
\usepackage{tikz}
\usepackage{comment}
\usepackage{amssymb} % define this before the line numbering.
\usepackage{siunitx}
\usepackage{color}
\usepackage[cmex10]{amsmath}
\usepackage{mathtools}
\usepackage{array}
\usepackage[utf8]{inputenc}
\usepackage{tabu}
\usepackage{tabularx}
\usepackage{booktabs}
\usepackage[english]{babel}
\usepackage{csquotes}
\usepackage[skip=2pt,font=small]{caption}
\usepackage{footnote}
\usepackage{amssymb}
\usepackage{verbatim}
\usepackage{multirow}
\usepackage{dsfont}
\usepackage{textcomp}
\usepackage{siunitx}
\usepackage{xcolor}
\usepackage{resizegather}
\usepackage{paralist}
\usepackage[mathscr]{euscript}
\usepackage{xfrac}
\usepackage{epsfig}
\usepackage{graphicx}
\usepackage{cite}
\definecolor{citecolor}{HTML}{0071bc}
\definecolor{linkcolor}{HTML}{e802af}
\definecolor{urlcolor}{HTML}{00a600}
\usepackage[pagebackref=false,breaklinks=true,colorlinks,citecolor=citecolor,linkcolor=linkcolor, urlcolor=urlcolor, bookmarks=false]{hyperref}

\usepackage[ruled,linesnumbered]{algorithm2e}

\usepackage[accsupp]{axessibility}  % Improves PDF readability for those with disabilities.

\begin{document}
% \linenumbers
\pagestyle{headings}
\mainmatter
\def\ECCVSubNumber{5777}  % Insert your submission number here

% \title{Generalizable Implicit Representation for Single-Shot Scene Reconstruction and Categorical 6D Pose and Size Estimation} % Replace with your title
% \title{GENIE: Generalizable Implicit Representations for 3D Autolabeling and Single-Shot Object Understanding}
% \title{GENIE: Generalizable Implicit Representations for Object Reconstruction}
\title{\titleLong}

%\begin{comment}
% \titlerunning{ECCV-22 submission ID \ECCVSubNumber} 
% \authorrunning{ECCV-22 submission ID \ECCVSubNumber} 
% \author{Muhammad Zubair Irshad*, Serg}
% \institute{Paper ID \ECCVSubNumber}
%\end{comment}
%******************

% \begin{comment}
\authorrunning{M.Z. Irshad, S. Zakharov et al.}
\titlerunning{ShAPO}

\newcommand{\orcid}[1]{\,\href{https://orcid.org/#1}{\protect\includegraphics[width=8pt]{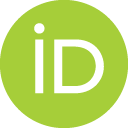}}}

\author{Muhammad Zubair Irshad*\inst{1}\orcid{0000-0002-1955-6194}\index{Irshad, Muhammad Zubair} \and
Sergey Zakharov*\inst{2}\orcid{0000-0002-6231-6137} \and
Rares Ambrus\inst{2}\orcid{0000-0002-3111-3812} \and
Thomas Kollar\inst{2}\orcid{0000-0003-2598-8118} \and
Zsolt Kira\inst{1}\orcid{0000-0002-2626-2004}\and
Adrien Gaidon\inst{2}\orcid{0000-0001-8820-550X}\\
\scriptsize{* denotes equal contribution}}
\institute{Georgia Institute of Technology  \and
Toyota Research Institute \\ \inst{1}\email{\{mirshad7, zkira\}@gatech.edu}, \inst{2}\email{\{first.last\}@tri.global}}

% \footnote{}


% \institute{Georgia Institute of Technology \\ \email{\{chenyun, mtimm, smaji\}@cs.umass.edu}}

% \author{Muhammad Zubair Irshad*\inst{1}\orcidID{0000-0002-1955-6194} \and
% Sergey Zakharov*\inst{2}\orcidID{1111-2222-3333-4444} \and
% Rares Ambrus\inst{3}\orcidID{2222--3333-4444-5555}}
% \end{comment}

%******************
\maketitle
\begin{sloppypar}
\input{sections/abstract.tex}
\input{sections/introduction.tex}
\input{sections/related_works}
\input{sections/method}
\input{sections/experiments}

\input{sections/qualitative}
\input{sections/conclusion}
\clearpage
\bibliographystyle{splncs04}
\bibliography{egbib}
\clearpage
\input{sections/appendix}

\end{sloppypar}
\end{document}

%% file: sections/abstract.tex
\begin{figure}
   \vspace{-0.5cm}
   \centering
%   \begin{center}
       % \fbox{\rule{0pt}{2in} \rule{.9\linewidth}{0pt}}
       \includegraphics[width=\linewidth]{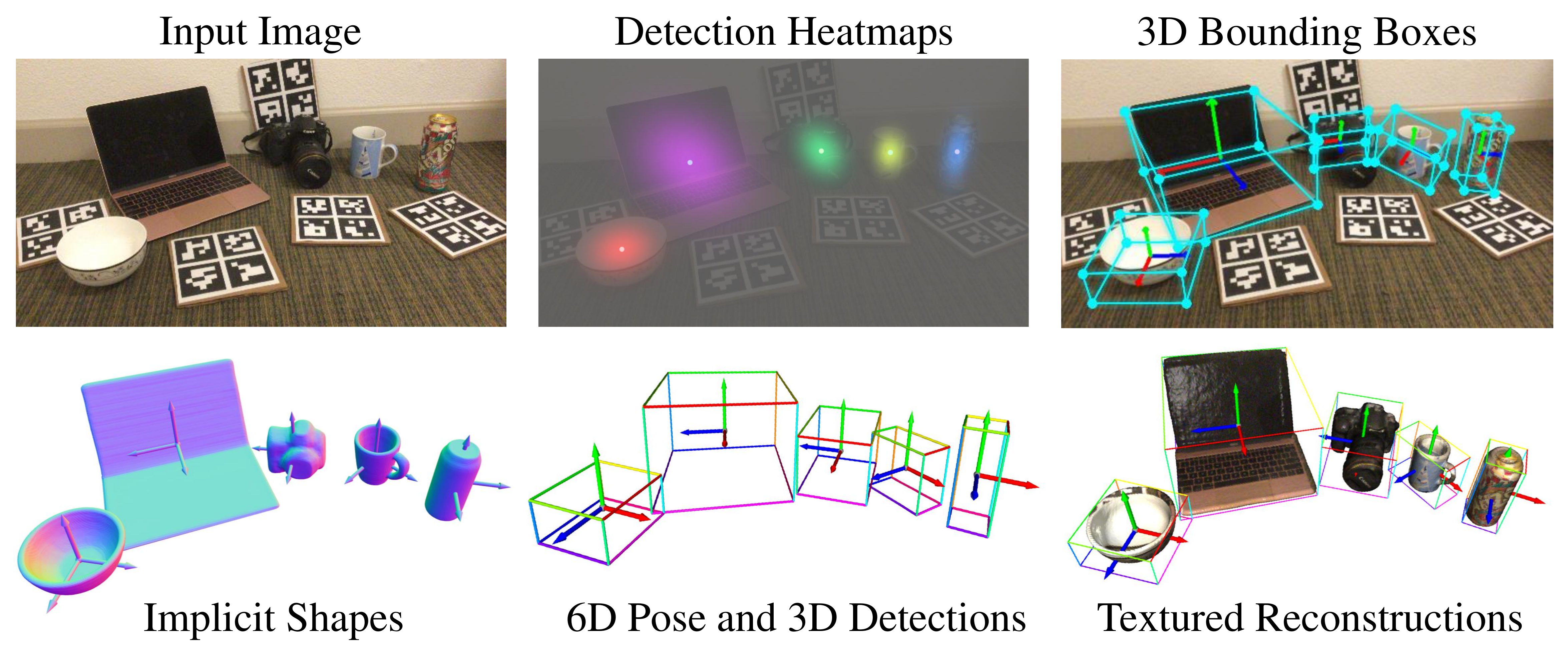}
%   \end{center}
%   \vspace{-0.75cm}
      \caption{Given a single RGB-D observation, our method, \textbf{ShAPO}, infers 6D pose and size, 3D shapes and appearance of all objects in the scene. Results shown on novel real-world scene in NOCS~\cite{wang2019normalized}.}
   \label{fig:intro}
   \vspace{-0.5cm}
\end{figure}

\looseness=-1
\begin{abstract}

% \zk{This seems to be a different list than Figure 1 (e.g. no texture). In general, good to use consistent terminology and enumerated list to avoid confusion (this is something that should be addressed throughout the paper)}
% We leverage implicit representations and differentiable rendering to

Our method studies the complex task of object-centric 3D understanding from a single RGB-D observation. As it is an ill-posed problem, existing methods suffer from low performance for both 3D shape and 6D pose and size estimation in complex multi-object scenarios with occlusions. We present ShAPO, a method for joint multi-object detection, 3D textured reconstruction, 6D object pose and size estimation.
Key to \titleShort
is a single-shot pipeline to regress shape, appearance and pose latent codes along with the masks of each object instance, which is then further refined in a sparse-to-dense fashion. A novel disentangled shape and appearance database of priors is first learned to embed objects in their respective shape and appearance space. We also propose a novel, octree-based differentiable optimization step, allowing us to further improve object shape, pose and appearance simultaneously under the learned latent space, in an analysis-by-synthesis fashion. Our novel joint implicit textured object representation allows us to accurately identify and reconstruct novel unseen objects without having access to their 3D meshes. 
Through extensive experiments, we show that our method, trained on simulated indoor scenes, accurately regresses the shape, appearance and pose of novel objects in the real-world with minimal fine-tuning. Our method significantly out-performs all baselines on the NOCS dataset with an 8\% absolute improvement in mAP for 6D pose estimation. Project page:
\href{https://zubair-irshad.github.io/projects/ShAPO.html}{zubair-irshad.github.io/projects/ShAPO.html}

\keywords{Implicit Representations; 3D Shape and Texture Reconstruction; 6D Pose Estimation; Octree-based Differentiable Optimization}
\end{abstract}

%% file: sections/introduction.tex
\section{Introduction}
Holistic 3D object understanding~\cite{zhang2021holistic,Nie_2020_CVPR} from a single RGB-D observation has remained a popular yet challenging problem in computer vision~\cite{He2017,zhou2019objects,gkioxari2019mesh} and robotics~\cite{jiang2021synergies,ferrari1992planning,laskey2021simnet,irshad2022centersnap}. The capability to infer complete object-centric 3D scene information has benefits in robotic manipulation~\cite{cifuentes2016probabilistic,jiang2021synergies, laskey2021simnet, irshad2022centersnap}, navigation~\cite{irshad2022sasra,9561806} and augmented reality. This task demands that the autonomous robot reasons about the 3D geometry of objects from partially observed single-view visual data, infer a 3D shape and 6D pose and size ( i.e. 3D orientation and position) and estimate the appearance of novel object instances. Despite recent progress, this problem remains challenging since inferring 3D shape from images is an ill-posed problem and predicting the 6D pose and 3D scale can be extremely ambiguous without having any prior information about the objects of interest. 

Prior works on object-centric scene understanding have attempted to address this challenge in various ways: \textit{object pose understanding} methods obtain 3D bounding box information without shape details. Most prior works in object-pose estimation cast it as an instance-level~\cite{kehl2017ssd,rad2017bb8,xiang2018posecnn} 3D object understanding task as opposed to category-level. Such methods~\cite{tekin2018real,peng2019pvnet,wang2019densefusion}, while achieving impressive results, rely on provided 3D reconstructions or prior CAD models for successful detection and pose estimation. Category-level approaches~\cite{tian2020shape,wang2019normalized,sundermeyer2020augmented}, on the other hand, rely on learned shape and scale priors during training, making them much more challenging. Despite great progress in category-level pose estimation, the performance of these approaches is limited due to their incapacity to express shape variations explicitly. 
\textit{Object-centric scene-level reconstruction} methods~\cite{wang2018pixel2mesh,niemeyer2020differentiable,kato2018} recover object shapes using 2D or partial 3D information for scene reconstruction. However, most methods are limited in their ability to reconstruct high quality shapes in a fast manner~(i.e. the studied representation is either voxel-based which is computationally inefficient~\cite{groueix2018} or point-based which results in poor reconstruction quality~\cite{fan2017point,choy20163d}).

For holistic scene-level reconstruction, only predicting shapes in an isolated manner~\cite{groueix2018,kuo2020mask2cad,yuan2018pcn} may not yield good results due to the challenges of aligning objects in the 3D space, reasoning about occlusions and diverse backgrounds. To the best our knowledge, fewer works have tackled the problem of joint shape reconstruction with appearance and object-centric scene context~(i.e. 3D bounding boxes and object pose and sizes) for a holistic object-centric scene understanding.

Motivated by the above, we present \titleLong, a learnable method unifying accurate shape prediction and alignment with object-centric scene context. As shown in Figure~\ref{fig:intro}, we infer the complete 3D information of novel object instances~(i.e. 3D shape along with appearance and 6D pose and sizes) from a single-view RGB-D observation; the results shown in Figure~\ref{fig:intro} are on a novel scene from the NOCS~\cite{wang2019normalized} dataset. In essence, our method represents object instances as center key-points~\cite{irshad2022centersnap,laskey2021simnet,duan2019centernet} in a spatial 2D grid. We regress the complete 3D information i.e. object shape and appearance codes along with the object masks and 6D pose and sizes at each objects' spatial center point. A novel joint implicit shape and appearance database of signed distance and texture field priors is utilized, to embed object instances in a unique space and learn from a large collection of CAD models. We further utilize differentiable optimization of implicit shape and appearance representation to iteratively improve shape, pose, size and appearance jointly in an analysis-by-synthesis fashion. To alleviate the sampling inefficiency inherent in signed distance field shape representations, we propose a novel octree-based point sampling which leads to significant time and memory improvements as well as increased reconstruction quality. 

Our contributions are summarized as follows:
\begin{itemize}
    \item{\textbf{A generalizable, disentangled shape and appearance space} coupled with \textbf{an efficient octree-based differentiable optimization procedure} which allows us to identify and reconstruct novel object instances without access to their ground truth meshes.}
    \item \textbf{An object-centric scene understanding pipeline} relying on learned joint appearance and implicit differentiable shape priors which achieves state-of-the-art reconstruction and pose estimation results on benchmark datasets.
    \item Our proposed approach significantly outperforms all baselines for 6D pose and size estimation on NOCS benchmark, showing over over 8\% absolute improvement in mAP for 6D pose at 10$^{\circ}$ \SI{10}{\cm}.
\end{itemize}

%% file: sections/related_works.tex
\section{Related Work}
In essence, our proposed method infers 3D shape along with predicting the 3D appearance and 6D pose and sizes of multiple object instances to perform object-centric scene reconstruction from a single-view RGB-D, so it relates to multiple areas in 3D scene reconstruction, object understanding and pose estimation.

\noindent\textbf{Neural Implicit Representations:}
3D shape and appearance reconstructions has recently seen a new prominent direction to use neural nets as scalar fields approximators instead of ConvNets. The first works of this class are notably DeepSDF~\cite{park2019deepsdf}, Occ-Net~\cite{mescheder2019occupancy}, and IM-Net~\cite{chen2019learning}. These works use \textit{coordinate based representation} to output either an occupancy estimate or a continuous SDF value, encoding the object's surface given a 3D coordinate. Improving this direction further, MeshSDF~\cite{remelli2020meshsdf}, NGLOD~\cite{takikawa2021neural} and Texture fields~\cite{OechsleICCV2019} employed end-to-end differentiable mesh representation, efficient octree representation and implicitly representing high frequency textures respectively. In our pipeline, we build a differentiable database of shape and appearance priors based on the latest advances in neural implicit representations. Our pipeline is end-to-end differentiable and abstains from the expensive Marching Cubes computation at every optimization step. Our database stores not only the geometries of the objects in the form of signed distance fields, but also their appearance in the form of texture fields, allowing us to model multiple categories through a single network while also considering test-time optimization of shape, pose and appearance.

\noindent\textbf{Object-centric Reconstruction:} 3D object reconstruction from a single-view observation has seen great progress to output pointclouds, voxels or meshes~\cite{park2019deepsdf,mescheder2019occupancy,chen2019learning}. Similarly, scene representation has been extended to include appearance. SRN~\cite{sitzmann2019scene}, DVR~\cite{niemeyer2020differentiable} learn from multi-view images by employing ray-marching routine and differentiable rendering respectively. Recently, NeRF~\cite{mildenhall2020nerf} propose to regress density and color along a ray and perform volumetric rendering to obtain true color value. Most NeRF-based methods~\cite{li2020neural, Niemeyer2020GIRAFFE, Ost_2021_CVPR} overfit to a single scene, do not promise generalizability and requires dense viewpoint annotations. Our pipeline, on the other hand, is capable of reconstructing shapes, appearances and inferring the 6D pose and sizes of objects never seen during training from a single-view RGB-D and is also not limited to viewpoints seen during training.

\noindent\textbf{6DoF Object Pose and size estimation} works use direct pose regression~\cite{irshad2022centersnap, kehl2017ssd, wang2019densefusion, xiang2018posecnn}, template matching~\cite{kehl2016deep, sundermeyer2018implicit, tejani2014latent} and establishing correspondences~\cite{wang2019normalized, dpodv2, park2019pix2pose,hodan2020epos, goodwin2022}. However, most works focus only on pose estimation and do not deal with shape and appearance retrieval and their connection to 6D pose. We instead propose a differentiable pipeline to improve the initially regressed pose, along with the shape and appearance, using our novel octree-base differentiable test-time optimization.   

%% file: sections/method.tex
\section{Method}
\titleShort is a learning-based holistic object-centric scene understanding method that infers 3D shape along with the 6D pose, size and appearance of multiple unknown objects in an R         GB-D observation. \titleShort tackles the detection, localization and reconstruction of all unknown object instances in 3D space. Such a goal is made possible by three components: 1) A single-shot detection and 3D prediction module that detects multiple objects based on their center points in the 2D spatial grid and recovers their complete 3D shapes, 6D pose and sizes along with appearance from partial observations. 2) An implicit joint differentiable database of shape and appearances priors which is used to embed objects in a unique space and represent shapes as signed distance fields~(SDF) and appearance as continuous texture fields~(TF). 3) A 2D/3D refinement method utilizing an octree-based coarse-to-fine differentiable optimization to improve shape, pose, size and appearance predictions iteratively.

\begin{figure}[t]
   \centering
       \includegraphics[width=\linewidth]{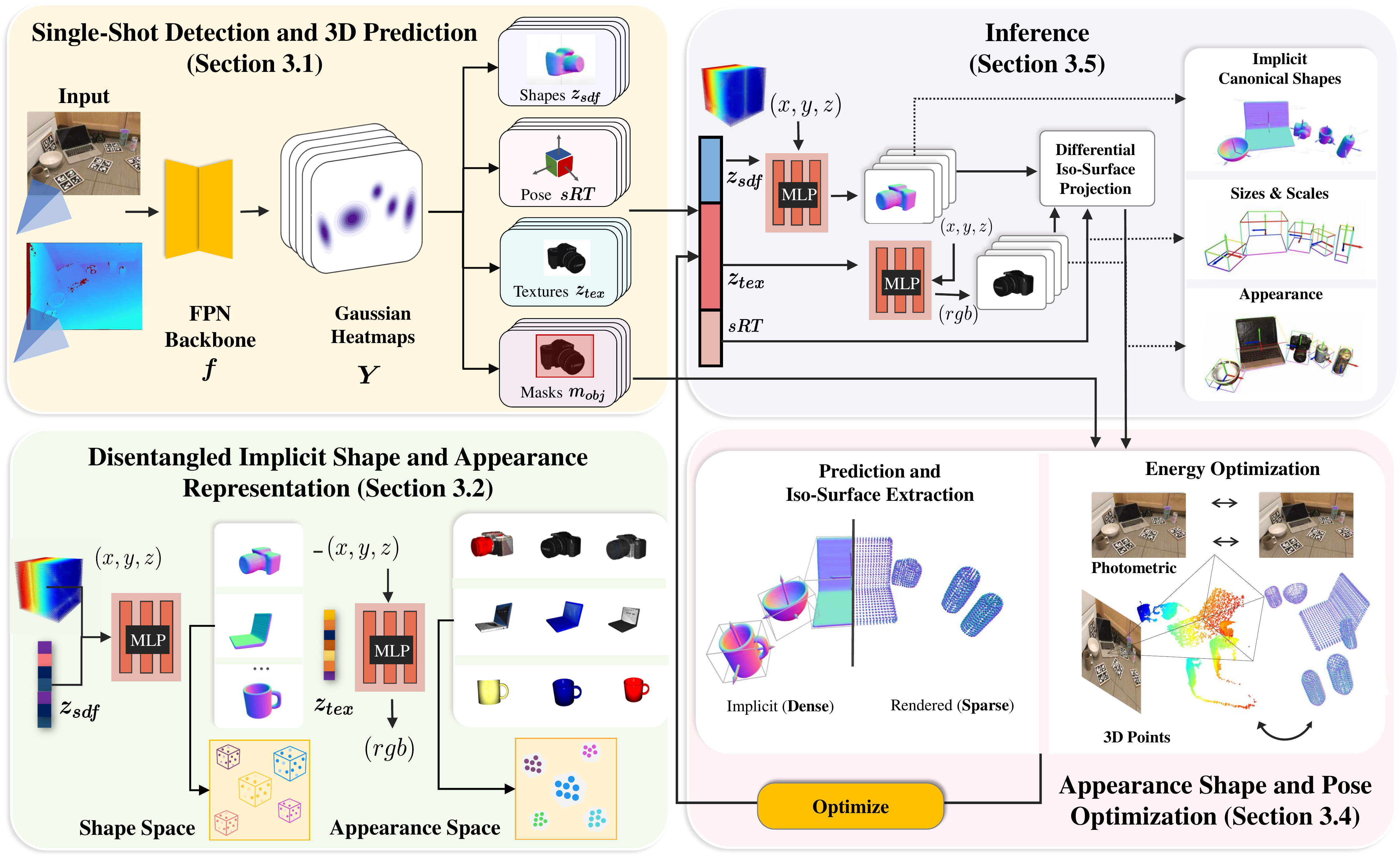}
      \caption{\textbf{\titleShort Method:} Given a single-view RGB-D observation, our method jointly predicts shape, pose, size and appearance codes along with the masks of each object instances in a single-shot manner. Leveraging a novel octree-based differentiable optimization procedure, we further optimize for shape, pose, size and appearance jointly in an analysis by synthesis manner. 
    }
   \label{fig:method}
\end{figure}

Formally, given a single-view RGB-D observation~($I$ $\in$ $\mathds{R}^{h_{o}\times w_{o}\times 3 }$, $D$ $\in$ $\mathds{R}^{h_{o}\times w_{o}}$) as input, \titleShort infers the complete 3D information of multiple objects including the shape~(represented as a zero-level set of an implicit signed distance field, $\boldsymbol{SDF}$), 6D pose and scales~($\tilde{\mathbf{\mathcal P}}$ $\in$ $SE(3)$, $\hat{s}$ $\in$ $\mathbb{R}^3$) and appearance (represented as a continuous texture field, $\boldsymbol{TF}$). To achieve the above goal, we employ a two-stage approach comprising of a single-shot inference
to extract initial pose, shape, and appearance estimates, and subsequently optimizing the latent codes and poses keeping network weights fixed. As shown in Figure~\ref{fig:method}, we first formulate object detection as a spatial per-pixel point detection~\cite{irshad2022centersnap,laskey2021simnet,zhou2019objects}. A backbone comprising of feature pyramid networks~\cite{girshick14CVPR} is employed with a specialized head to predict object instances as heatmaps in a per-pixel manner. Second, joint shape, pose, size and appearance codes along with instance masks are predicted for detected object centers using specialized heads~(Section~\ref{method:section1}). Our combined shape and appearance implicit differentiable database of priors is described in Section~\ref{method:section2} and the corresponding zero iso-surface based
differentiable rendering is detailed in Section~\ref{method:section3}. Lastly, 3D shapes along with their appearance, coupled with 6D pose and sizes of novel objects are jointly optimized during inference to predict accurate shapes, poses and sizes of novel object instances~(Section~\ref{method:section4}).

\subsection{Single-Shot Detection and 3D prediction}\label{method:section1}

\titleShort represents object instances along with their complete 3D information including shape, pose, appearance and size codes along with corresponding 2D masks through their 2D location in the spatial RGB image, following~\cite{irshad2022centersnap,laskey2021simnet,zhou2019objects,duan2019centernet}. Given an RGB-D observation~($I$ $\in$ $\mathds{R}^{h_{o}\times w_{o}\times 3 }$, $D$ $\in$ $\mathds{R}^{h_{o}\times w_{o}}$), \titleShort predicts object-centric heat maps~$\hat{Y}\in [0,1]^{\frac{h_{o}}{R} \times \frac{w_{o}}{R}\times 1}$ where each detected point~$(\hat{x}_{i}, \hat{y}_{i})$ denotes the local maxima in the heatmap~($\hat{Y}$).
% \zk{How do you determine how many heatmaps to output?}
Here $R$ denotes the heatmap down-sampling factor, and is set to $8$ in all our experiments. To predict these heatmaps, a feature backbone based on feature pyramid networks (FPN)~\cite{kirillov2019panoptic} is utilized along with a specialized heatmap prediction head.
During training, we find the target heatmaps by splatting the ground truth center points~$(x_{i}, y_{i})$ using the gaussian kernel~$\mathcal{N}(x_{i}, y_{i}, \sigma_{i})$ where $\sigma_{i}$ is relative to the spatial extent of each object (as defined by the corresponding ground truth bounding box annotation).
The network is trained to predict ground-truth heatmaps~($Y$) by minimizing MSE loss over all pixels $(x,y)$ in the heatmap, $\mathcal{L}_{\text{inst}}=\sum_{xyg}\left(\hat{Y}- Y\right)^{2}$. The network also predicts object instance masks~($\hat{M}$) using a specialized head~($f_{\theta_{m}}$) to output $\hat{M} \in \mathds{R}^{h_{o}\times w_{o}}$, similar to the semantic segmentation head described in~\cite{kirillov2019panoptic}. Note that it is crucial for the network to predict masks for accurate downstream optimization~(see Section~\ref{method:section3}).

\subsection{Joint Implicit Shape, Pose and Appearance Prediction}\label{method:section2}
Once the network detects objects, it then predicts their complete 3D information~(i.e. 3D shape, 6D pose and size along with the 3D appearance) all in a single-forward pass using specialized heads~($f_{\theta_{sdf}}$, $f_{\theta_{P}}$ and $f_{\theta_{tex}}$) with outputs~($Y_{sdf} \in \mathds{R}^{\frac{h_{o}}{R}\times \frac{w_{o}}{R}\times 64}$, $Y_{P}\in \mathds{R}^{\frac{h_{o}}{R}\times \frac{w_{o}}{R}\times 13}$ and $Y_{tex}\in \mathds{R}^{\frac{h_{o}}{R}\times \frac{w_{o}}{R}\times 64}$ respectively).
During training, the task-specific heads output shape code~$z_{sdf}$, 6D pose~$\tilde{\mathbf{\mathcal P}}$, scale~$\hat{s}$ and appearance~$z_{tex}$ information at each pixel in the down-sampled map~($\frac{h_{o}}{R}\times \frac{w_{o}}{R}$).
For each object's Pose~($\tilde{\mathbf{\mathcal P}}$) with respect to the camera coordinate frame, we regress a 3D rotation $\hat{\mathcal{R}}$ $\in$ $SO(3)$, a 3D translation $\hat{t}$ $\in$ $\mathbb{R}^3$ and 1D scales~$\hat{s}$ (totaling thirteen numbers). These parameters are used to transform the object shape from canonical frame to the 3D world. We select a 9D rotation~$\hat{\mathcal{R}}$ $\in$ $SO(3)$ representation since the neural network can better fit a continuous representation and to avoid discontinuities with lower rotation dimensions, as noted in~\cite{zhou2019continuity}. Furthermore, we employ a rotation mapping function following~\cite{pitteri2019object} to handle ambiguities caused by rotational symmetries. The rotation mapping function is used only for for symmetric objects~\textit{(bottle, bowl, and can)} in our database during training and it maps ambiguous ground-truth rotations to a single canonical rotation by normalizing the pose rotation.
Note that during training, ground-truth shape codes~$z_{sdf}$ and appearance codes~$z_{tex}$ for each object are obtained from our novel implicit textured differentiable representation~(further described in Section~\ref{shapeappearanceemb}).

During training, we jointly optimize for shape, pose, appearance and mask prediction. Specifically, we minimize the masked $L_{1}$ loss for shape, pose and appearance prediction, denoted as $\mathcal{L}_{\text{sdf}},  \mathcal{L}_{\text{tex}},  \mathcal{L}_{\text{P}}$ and a pixel-wise cross-entropy loss for mask prediction $ \mathcal{L}_{\text {M}}=\sum_{i=1}^{h_{o} \cdot w_{o}}-\log \hat{M}_{i}\left(M_{i}^{\mathrm{gt}}\right)$ where $M_{i}^{\mathrm{gt}}$ denotes the ground truth category label for each pixel.

During training, we minimize a combination of these losses as follows:
\begin{equation}
\mathcal{L}= \lambda_{inst}\mathcal{L}_{inst} + \lambda_{sdf}\mathcal{L}_{sdf} +  \lambda_{tex}\mathcal{L}_{tex} + \lambda_{M}\mathcal{L}_{M} + \lambda_{P}\mathcal{L}_{P}
\end{equation}
where $\lambda$ is a weighting co-efficient with values determined empirically as $lambda_{inst} = 100$ and $\lambda_{sdf} =  \lambda_{tex} =  \lambda_{P}$ = 1.0.
Note that for shape, appearance and pose predictions, we enforce the $L_{1}$ loss based on the probability estimates of the Gaussian heatmaps ($Y$) i.e. the loss is only applied where Y has a score greater than 0.3
to prevent ambiguity in the space where no object exists. We now describe the shape and appearance representation utilized by our method. 
\subsection{Implicit textured differentiable database of priors}\label{shapeappearanceemb}
\begin{figure}[t]
   \centering
       \includegraphics[width=\linewidth]{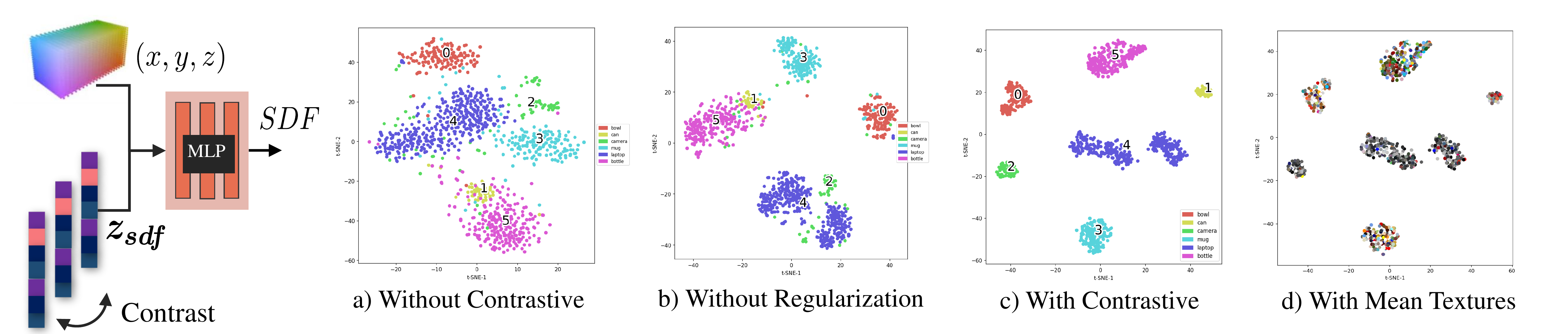}
      \caption{Deep Implicit shape space for NOCS object Categories}
   \label{fig:shape_space}
\end{figure}

We propose a novel joint implicit textured representation to learn from a large variety of CAD models and embed objects in a concise space. This novel representation~(as shown in Figure~\ref{fig:method} and Figure~\ref{fig:shape_space}) is also used as a strong inductive prior to efficiently optimize the shape and appearance along with the pose and size of objects in a differentiable manner~(Section~\ref{method:section3}). In our implicit textured database of shape and appearance priors, each object shape is represented as a Signed Distance Field~(SDF) where a neural network learns a signed distance function $G(\boldsymbol{x},\boldsymbol{z}_{sdf})=s: z_{sdf} \in \mathbb{R}^{64} , s\in \mathbb{R}$ for every 3D point $x \in \mathbb{R}^{3}$ and the appearance is represented as Texture Fields~($t_{\theta}: \mathbb{R}^{3} \rightarrow \mathbb{R}^{3}$) which maps a  3D point $x \in \mathbb{R}^{3}$ to an RGB value $c \in \mathbb{R}^{3}$. Since the mapping between coordinates and colors is ambiguous without shape information, we propose to learn a texture field only at the predicted shape i.e. $t_{\theta}(\boldsymbol{x},\boldsymbol{z}_{sdf},\boldsymbol{z}_{tex})=c,\boldsymbol{z}_{tex} \in \mathbb{R}^{64}$. The SDF function~($G$) implicitly defines the surface of each object shape by the zero-level set $G(.) = 0$. To learn a shape-code~($\boldsymbol{z}_{sdf}$) and texture code~($\boldsymbol{z}_{tex}$) for each object instance, we design a single MLP~(multi-layer perceptron) each for shape (to reason about the different geometries in the database) and texture~(to predict color information given shape and texture codes). Through conditioning the MLP output on the latent vector, we allow modeling multiple geometries and appearances using a single network each for shape and appearance. Each MLP is trained separately using the supervised ground-truth reconstruction loss $L_{SDF}$ and $L_{RGB}$ as follows:

\begin{gather}
    L_{SDF}=\left|\operatorname{clamp}\left(G(\boldsymbol{x},\boldsymbol{z}_{sdf}), \delta\right)-\operatorname{clamp}\left(\boldsymbol{s}_{gt}, \delta\right)\right| + L_{contrastive}(\boldsymbol{z}_{sdf}) \\
L_{RGB} = \sum_{n=1}^{N} \|\boldsymbol{c}_{gt}-t_{\theta}(\boldsymbol{x},\boldsymbol{z}_{sdf},  \boldsymbol{z}_{tex})\|_{2}^{2}
\end{gather}  

where $L_{SDF}$ is a combination of a clipped $L_{1}$ loss between ground-truth signed-distance values $\boldsymbol{s}_{gt}$ and predicted SDF $G(\boldsymbol{x},\boldsymbol{z}_{sdf})$ and a contrastive loss~$L_{\text {contrastive }}=\left[m_{\text {pos  }}-s_{p}\right]_{+}+\left[s_{n}-m_{\text {neg }}\right]_{+}$. As shown by the t-SNE embeddings~\cite{van2008visualizing} (Figure~\ref{fig:shape_space}) for the latent shape-code~($\boldsymbol{z}_{sdf}$), the contrastive loss helps with disentangling the shape space nicely and leads to better downstream regression in the single-shot model (Section ~\ref{method:section2}). Once we train the implicit shape auto-decoder, we use the learned shape space $\boldsymbol{z}_{sdf}$ to minimize the color loss $L_{RGB}$ which is defined as an MSE loss between predicted color at the surface~$t_{\theta}(\boldsymbol{x},\boldsymbol{z}_{sdf},\boldsymbol{z}_{tex})$ and ground-truth color~$\boldsymbol{c}_{gt}$.

We use 3D textured models from the CAD model repository Shapenet~\cite{chang2015shapenet} to learn our database of shape and texture priors. Once trained, the MLP networks for both shape and appearance find a disentangled space for color and geometry while keeping semantically-similar objects together~(Figure~\ref{fig:shape_space}) and provides us with strong priors to be used for 2D and 3D optimization~(described in Section ~\ref{method:section3}).

\subsection{Differentiable Optimization}\label{method:section3}
A key component of \titleShort is the optimization scheme allowing to refine initial object predictions with respect to the pose, scale, shape, and appearance. Inspired by sdflabel~\cite{zakharov2020autolabeling}, we develop a new differentiable and fast optimization method. Instead of using mesh-based representations, we rely entirely on implicit surfaces, which not only helps us avoid common connectivity and intersection problems, but also provides us full control over sampling density. 
\begin{figure}[b]
   \centering
       \includegraphics[width=\linewidth]{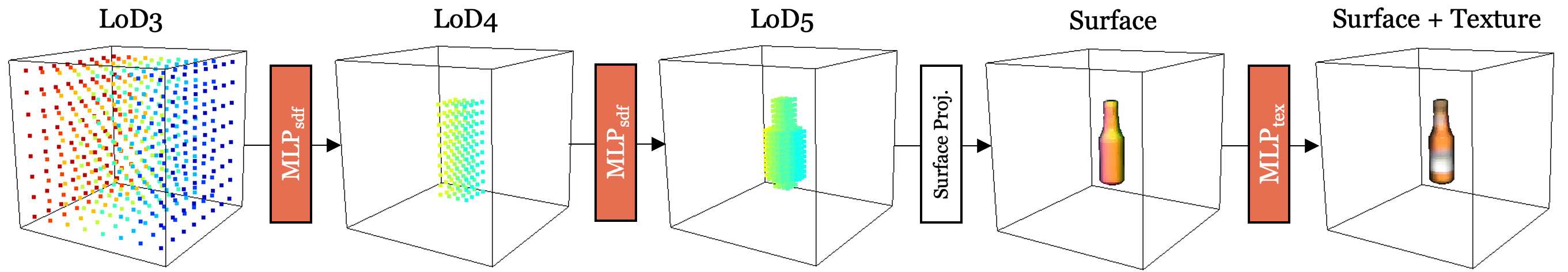}
      \caption{Octree-based object extraction.}
   \label{fig:octree}
\end{figure}

\looseness=-1
\subsubsection{Surface Projection}
Given input grid points $x_i$ and estimated SDF values $s_i$, we aim to find a differentiable transformation to extract the object surface encoded in $\mathbf{z}_{sdf}$. A trivial solution would be to simply threshold points with SDF values more than a certain threshold. However, this procedure is not differentiable with respect to the input latent vector $\mathbf{z}_{sdf}$. Instead, we utilize the fact that deriving an SDF value $s_i$ with respect to its input coordinate $x_i$ yields a normal at this point, which can be computed in a single backward pass:

\begin{equation}
    n_i = \frac{\partial G(x_i;\mathbf{z}_{sdf})}{\partial x_i}.
\quad\text{and}\quad
    p_i = x_i - \frac{\partial G(x_i;\mathbf{z}_{sdf})}{\partial x_i} G(x_i;\mathbf{z}_{sdf}).
\end{equation}

\subsubsection{Octree-based Point Sampling}
The brute force solution to recover shapes from a learned SDF representation can be obtained by estimating SDF values for a large collection of grid points similar to the procedure used in \cite{zakharov2020autolabeling}. To obtain clean surface projections one would then disregard all points $x_i$ outside a narrow band ($|s_i| > 0.03$) of the surface. However, this procedure is extremely inefficient both memory- and compute-wise --- For a grid size of $60^3 = 216000$ points, only around 1600 surface points are extracted (accounting to $0.7\%$ of the total number of points). We propose an Octree-based procedure to efficiently extract points.
We define a coarse voxel grid and estimate SDF values for each of the points using our trained SDF network. We then disregard voxels whose SDF values are larger than the voxel grid size for this resolution level. The remaining voxels are subdivided each generating eight new voxels. We repeat this procedure until the desired resolution level is reached. In our implementation, we start from level of detail (LoD) 3 and traverse up to LoD 6 depending on the desired resolution level. Finally, when points are extracted we estimate their SDF values and normals and project them them onto the object surface. The pseudo-code implementation of the Octree-based surface extraction is provided in Alg.~\ref{alg:octree} with the visualization shown in Fig.~\ref{fig:octree}.

\input{tables/optimization}
\vspace{-4mm}

\looseness=-1
\subsection{Inference:}\label{method:section4}
During inference, we first perform predictions using our single-shot model. Object detection is performed using peak detection~\cite{zhou2019objects} on the outputs of predicted heatmaps~($\hat{Y}$). Each detected center point~{($\boldsymbol{x}_{i}, {\boldsymbol{y}_{i}}$)} corresponds to maxima in the heatmap output~($\hat{Y}$). Second, we sample shape, pose and appearance codes of each object from the output of task-specific heads at the detected center location~($x_{i}, {y_{i}}$) via~$\boldsymbol{z}_{sdf} = Y_{sdf}(\boldsymbol{x}_{i}, {\boldsymbol{y}_{i}})$, ~$\boldsymbol{z}_{tex} = Y_{tex}(\boldsymbol{x}_{i}, {\boldsymbol{y}_{i}})$ and ~$\tilde{\mathbf{\mathcal P}} = Y_{P}(\boldsymbol{x}_{i}, {\boldsymbol{y}_{i}})$.
We utilize the predicted shape, pose, size and appearance codes as an initial estimate to further refine through our differentiable optimization pipeline. Our optimizer takes as input the predicted implicit shapes in the canonical frame of reference along with the masks predictions~($\hat{M}$),  color codes~($\boldsymbol{z}_{tex}$) and extracted~$3\times 3$ rotation $\hat{\mathcal{R}}_{i}^{p}$, 3D translation vector $\hat{t}_{i}^{p}$ and 1D scales $\hat{s}_{i}^{p}$ from recovered Pose $\tilde{\mathbf{\mathcal P}}$. Although a variety of related works consider mean class predictions as initial priors, we mainly utilize the regressed outputs of shape, appearance and pose for the optimization pipeline since the initial estimates are very robust~(see Table~\ref{comparison_table}). We utilize the predicted SDF to recover the complete surface of each object, in a coarse-to-fine manner, using the proposed differentiable zero-isosurface projection~(Section ~\ref{method:section3}). After fixing the decoder~(G) parameters, we optimize the feature vector $\boldsymbol{z}_{sdf}$ by estimating the nearest neighbour between the predicted projected pointclouds and masked pointclouds obtained from depth map and predicted masks~($\hat{M}$) of each object. In essence, a shape code $\boldsymbol{z}_{sdf}$ is refined using the Maximum-a-Posterior (MAP) estimation as follows:
\begin{equation}
\boldsymbol{z}_{sdf}=\underset{\boldsymbol{z}}{\arg \min } (\mathcal{L}\left(D(G(\boldsymbol{z}, \boldsymbol{x}\right)), P_{d})
\end{equation}
where D() denotes the differentiable iso-surface projection described in Section~\ref{method:section3} and $P_{d}$ denotes the pointclouds obtained from masked depth maps. We further optimize the RGB component similarly by optimizing the difference in colors between the masked image color values~($C_{d}$) and colors obtained using the regressed color codes decoded by the texture field~($t_{\theta}$)  $\boldsymbol{z}_{tex}=\underset{\boldsymbol{z}}{\arg \min } (\mathcal{L}\left(D(t_{\theta}(\boldsymbol{z}, \boldsymbol{x}\right)), C_{d})$. We further allow $t_{\theta}$ weights to change to allow for a finer level of reconstruction. For qualitative results please consult supplementary material.

%% file: tables/optimization.tex
\looseness=-1
\begin{algorithm}[t!]
	\KwIn{ 
			$\textbf{x} \in \mathbb{R}^3$ grid points,
			$\textbf{l} \in \mathbb{L}$ grid levels,
			$\textbf{z}_{sdf}$ and $\textbf{z}_{tex} \in \mathbb{R}^{64}$ latent vectors 
		}
		\KwOut{ 
			$\textbf{pcd} \in \mathbb{R}^3$ surface points,
			$\textbf{nrm} \in \mathbb{N}^3$ normals,
			$\textbf{col} \in \mathbb{C}^3$ colors
		}
		
		\tcc{Extract object grid (no grad)}
% 		$sum = 0$ \;
		\For{$ l \in \{1, \ldots, N_{LoD}\} $}{
            $\textbf{sdf} \leftarrow G(\textbf{x}_l, \textbf{z}_{sdf})$ \tcp*{regress $sdf$ values}
            $\textbf{occ} \leftarrow \textbf{sdf} < getCellSize(l)$ \tcp*{estimate cell occupancy}
			$\textbf{x}_{l_{occ}} \leftarrow \textbf{x}_l[occ]$ \tcp*{remove unoccupied cells}
			$\textbf{x}_{l+1} \leftarrow subdivide(\textbf{x}_{l_{occ}})$ \tcp*{subdivide cells to go to next LoD}
			
		}
		\tcc{Extract object shape and appearance}
		%	$p \leftarrow \{\}$ \;
		$\textbf{nrm} \leftarrow \text{backprop}(\textbf{sdf})$ \tcp*{analytically estimate surface normals}
		$\textbf{pcd} \leftarrow \textbf{x} - \textbf{nrm} * \textbf{sdf}$ \tcp*{project points onto the surface}
		$\textbf{col} \leftarrow t_{\theta}(\textbf{pcd}, \textbf{z}_{sdf}, \textbf{z}_{tex})$ \tcp*{regress surface texture}
		\KwRet{$\textbf{pcd}$, $\textbf{nrm}$, $\textbf{col}$}
		
		\caption{Octree-based implicit surface extraction}
		\label{alg:octree}
	\end{algorithm}

%% file: sections/experiments.tex
\input{tables/comparison}
\section{Experiments}
In this section, we empirically validate the peformance of our method. In essence, our goal is to answer these questions:
1) How well does ShAPO recover pose and sizes of novel objects.
2) How well does ShAPO perform in terms of reconstructing geometry and appearance of multiple objects from a single-view RGB-D observation? 3) How well does does our differentiable iterative improvement and multi-level optimization impact shape, appearance, pose and size?\\
\textbf{Datasets} We utilize the \textbf{NOCS}~\cite{wang2019normalized} dataset for both shape reconstruction and category-level 6D pose and size estimation evaluation. For training, we utilize the CAMERA dataset which comprises 300K synthetic images, of which 25K are hold-out for evaluation. The training dataset includes 1085 object object models from 6 different categories - \textit{bottle, bowl, camera, can, laptop and mug} whereas the evaluation dataset includes 184 different models. The REAL dataset train-set comprises of 7 scenes with 4300 images, and test-set comprises of 6 scenes with 2750 real-world images. \\
\looseness=-1
\textbf{Metrics}
We evaluate the the performance of 3D object detection and 6D pose estimation independently following~\cite{wang2019normalized}. We use the following key metrics: 1) Average-precision for different IOU-overlap thresholds (\textbf{IOU25} and \textbf{IOU50}). 2) Average precision for which the error is less than $n^{\circ}$ for rotation and $m$ cm for translation (\textbf{5\textdegree \SI{5}{\cm}}, \textbf{\textbf{5\textdegree \SI{10}{\cm}}} and \textbf{\textbf{10\textdegree \SI{10}{\cm}}}). We use Chamfer distance~(CD) for shape reconstruction following~\cite{yuan2018pcn}.
\begin{figure}[b]
  \centering
      \includegraphics[width=0.75\textwidth]{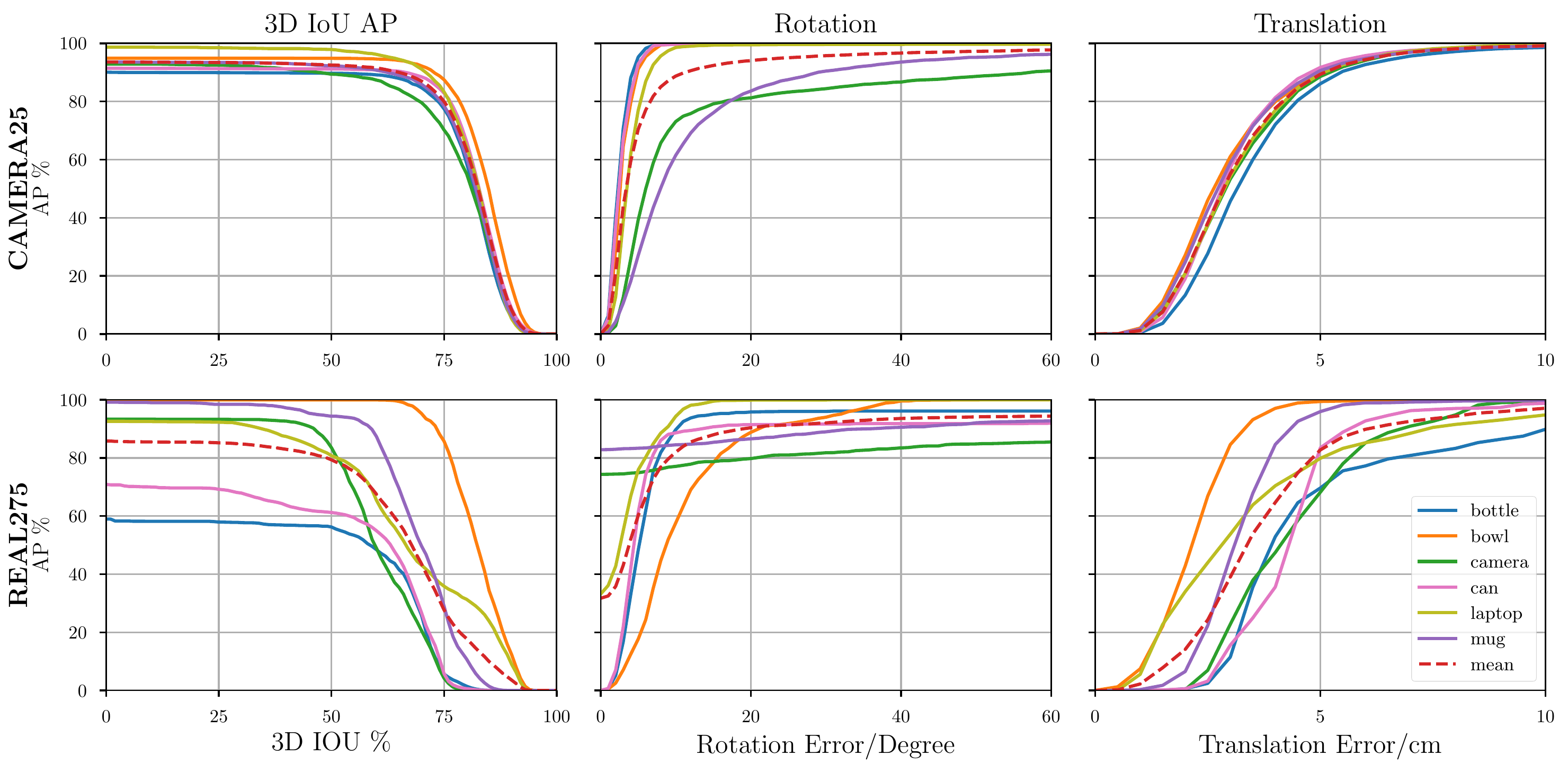}
      \caption{Average precision of ShAPO for various IOU, rotation and translation thresholds on NOCS CAMERA25 and REAL275 dataset.}
  \label{fig:mAP}
\end{figure}

\input{tables/reconstruction}
\subsection{Implementation Details}
\looseness=-1
\titleShort is sequentially trained first on the CAMERA set with minimal fine-tuning on Real training set. For SDF, we use an MLP with 8 layers and hidden size of 512. For color, we utilize a Siren MLP~\cite{sitzmann2020implicit} as it can fit higher frequencies better. We train the SDF and Color MLPs on all categories for 2000 epochs. We use Pytorch~\cite{NEURIPS2019_9015} for our models and training pipeline implementation. For optimization, we use an adaptive learning rate which varies with the obtained masks of each object since we believe masks capture the confidence of heatmap prediction during detection. We optimize each object for 200 iterations.

\subsection{NOCS 6D Pose and Size Estimation Benchmark}
\textbf{NOCS Baselines:} We compare eight model variants to show effectiveness of our method:
(1) \textbf{NOCS}~\cite{wang2019normalized}: Regresses NOCS map and predicts poses using similarity transform with depth maps. We report the best pose estimation configuration in NOCS~(i.e. 32-bin classification) (2) \textbf{Shape Prior}~\cite{tian2020shape}: Estimates shape deformation for inter-class variation. (3) \textbf{CASS}~\cite{chen2020learning}: Regress the pose and size with first detecting 2D bounding boxes. (4) \textbf{Metric-Scale}~\cite{lee2021category}: Estimates the object center and metric scale (5) \textbf{CenterSnap~\cite{irshad2022centersnap}:} Single-shot approach to predict pose and size (6) \textbf{CenterSnap-R~\cite{irshad2022centersnap}:} ICP optimization done with the predicted point-cloud based shape. Following~\cite{irshad2022centersnap}, we do not compare against 6D pose tracking baselines such as~\cite{wang20206,wen2021bundletrack} which need pose initializations and are not end-to-end detection based~(i.e. they do not report mAP metrics) \\
\textbf{Comparison with strong baselines on NOCS:} Table~\ref{comparison_table} and Figure~\ref{fig:mAP} shows the result of our proposed \titleShort method. \titleShort consistently outperforms all the baseline methods on 6D pose and size estimation and 3D object detection. Specifically, \titleShort method shows superior generalization on the REAL test-set by achieving a mAP of 85.3$\%$ for 3D IOU at 0.25, 57.0$\%$ for 6D pose at 5\textdegree \SI{10}{\cm} and 78.0$\%$ for 6D pose at 10\textdegree \SI{10}{\cm}, hence demonstrating an absolute improvement of 1.8$\%$, 25.4$\%$ and 7.1$\%$ over the best-performing baseline on the Real dataset. Our method also achieves better generalization on the never-seen CAMERA evaluation set. We achieve a mAP of 94.5$\%$ for 3D IOU at 0.25, 75.9$\%$ for 6D pose at 5\textdegree \SI{10}{\cm} and 89.2$\%$ for 6D pose at 10\textdegree \SI{10}{\cm}, demonstrating an absolute improvement of 1.3$\%$, 4.2$\%$ and 1.3$\%$ over the best-performing baseline.

\looseness=-1
\subsection{Generalizable Implicit Object Representation}
\begin{figure}[b!]
   \centering
       \includegraphics[width=\linewidth]{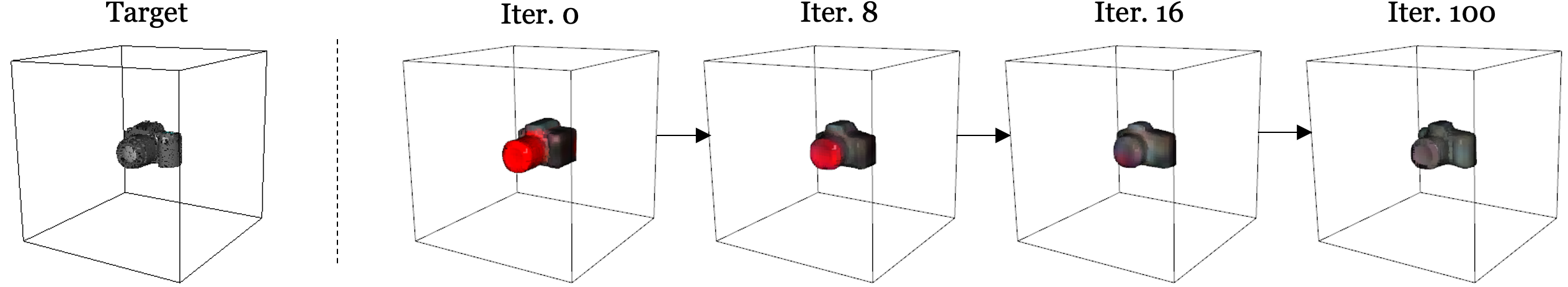}
      \caption{Shape/Appearance optimization.}
   \label{fig:optimization}
\end{figure}

In this experiment, we evaluate the effectiveness of our generalizable implicit object representation as well as the efficiency of our octree-based differentiable optimization. To do that, we isolate our implicit representation from the detection pipeline and consider the 3D reconstruction / identification task - given a novel object from NOCS test split, we optimize our representation for 200 iterations, while keeping $f_{sdf}$ and $f_{rgb}$ weights frozen, to find the best fitting model in terms of both shape and texture. We initialize the optimization using the average latent feature per class. The standard Adam solver with the learning rate of 0.001 for both for both shape and appearance losses (L2 norms) is used with weight factors 1 and 0.3 respectively. We use three different octree resolution levels - from LoD5 to LoD7. Additionally, we show three different resolution levels for the standard ordinary grid sampling~\cite{zakharov2020autolabeling} (40, 50, 60). Table~\ref{tab:optimization_3d} summarizes the results of the ablation by comparing different modalities with respect to the average point sampling (input vs output) and time efficiency, average GPU memory consumption, as well as reconstruction for shape (Chamfer distance) and texture (PSNR). One can see that our representation is significantly more efficient than the ordinary grid representation with respect to all metrics. While LoD7 provides best overall results, we use LoD6 for our experiments since it results in the optimal speed/memory/reconstruction trade-off. We show an example optimization procedure in Fig.~\ref{fig:optimization}.

\looseness=-1
\subsection{NOCS Object Reconstruction Benchmark}
\input{tables/optimization_3d}
To quantitatively analyze the reconstruction accuracy, we measure the Chamfer distance~(CD) between our reconstructed pointclouds and the ground-truth CAD model in NOCS. Our results are reported in Table~\ref{reconstruction_nocs}. Our results show consistently lower CD metrics for all class categories which shows superior reconstruction performance on novel object instances. We report a lower mean Chamfer distance of 0.14 on CAMERA25 and 0.15 on REAL275 compared to 0.20 and 0.32 reported by the competitive baseline~\cite{tian2020shape}.

%% file: tables/comparison.tex
\setlength{\tabcolsep}{1pt}
\begin{table}[t]
    \small
    \centering
    \renewcommand{\arraystretch}{1.3}
    \caption{
    \textbf{Quantitative comparison of 6D pose estimation and 3D object detection on NOCS}~\cite{wang2019normalized}: Comparison with strong baselines. Best results are highlighted in \textbf{bold}. $*$ denotes the method does not report IOU metrics since size and scale is not evaluated. We report metrics using nocs-level class predictions for a fair comparison with all baselines.
    }
    \label{comparison_table}
    % \vspace{+0.15cm}
    \resizebox{1.0\textwidth}{!}{
    \begin{tabular}{clcccccccccccc}
        \toprule
        & & \multicolumn{6}{c}{\textbf{CAMERA25}} & \multicolumn{6}{c}{\textbf{REAL275}} \\ 
        \cmidrule(r{0.1in}){3-8} \cmidrule(r{0.1in}){9-14}
        & {Method} & \textbf{IOU25} & \textbf{IOU50} & \textbf{5\textdegree \SI{5}{\cm}} & \textbf{5\textdegree \SI{10}{\cm}} & \textbf{10\textdegree \SI{5}{\cm}} & \textbf{10\textdegree \SI{10}{\cm}} & \textbf{IOU25} & \textbf{IOU50} & \textbf{5\textdegree \SI{5}{\cm}} & \textbf{5\textdegree \SI{10}{\cm}}& \textbf{10\textdegree \SI{5}{\cm}} & \textbf{10\textdegree \SI{10}{\cm}}\\
        \cmidrule(r{0.1in}){2-2}
        \cmidrule(r{0.1in}){3-8} \cmidrule(r{0.1in}){9-14}
        1 & {NOCS~\cite{wang2019normalized}}      &  91.1& 83.9 & 40.9  & 38.6 & 64.6 & 65.1 & 84.8 & 78.0 & 10.0 & 9.8 & 25.2 & 25.8\\
        2 & {Synthesis$^{*}$~\cite{chen2020category}} &  - & - & -  &- & - & - & - & - &  0.9 & 1.4 & 2.4 & 5.5 \\

        3 & {Metric Scale~\cite{lee2021category}}      & 93.8& 90.7 & 20.2  & 28.2 & 55.4 & 58.9 & 81.6 & 68.1 & 5.3 & 5.5 & 24.7 & 26.5 \\
        4 & {ShapePrior~\cite{tian2020shape}} &81.6	&72.4&	59.0&	59.6&  81.0 &  81.3 &	81.2& 77.3	&	21.4	&21.4&	54.1&	54.1\\
        5 & {CASS~\cite{chen2020learning}} & - & - & - & - & - & - & 84.2 & 77.7 &  23.5 & 23.8 & 58.0 & 58.3\\
        6 & {CenterSnap~\cite{irshad2022centersnap}} & 93.2&	92.3&	63.0	& 69.5 &	79.5 & 87.9&	83.5 &	80.2 & 27.2 &	29.2 & 58.8 &	64.4 \\
        7 & {CenterSnap-R~\cite{irshad2022centersnap}} & 93.2&	92.5&	66.2	& 71.7 &81.3 & 87.9&	83.5 &	\textbf{80.2} & 29.1 &	31.6 & 64.3 &	70.9 \\
       \midrule
        8 & {\textbf{\titleShort (Ours)}} & \textbf{94.5}&	\textbf{93.5}&	\textbf{66.6}	& \textbf{75.9} &	\textbf{81.9} & \textbf{89.2}&	\textbf{85.3} &	79.0 & \textbf{48.8} &	\textbf{57.0} & \textbf{66.8} &	\textbf{78.0} \\
      
        \bottomrule
    \end{tabular}
    }
    % \vspace{-2mm}
\end{table}

%% file: tables/reconstruction.tex
\setlength{\tabcolsep}{1pt}
\begin{table*}[t]
    \centering
    \small
    \renewcommand{\arraystretch}{1.3}
    % \vspace{-1mm}
    \caption{
    \textbf{Quantitative comparison of 3D shape reconstruction on NOCS}~\cite{wang2019normalized}: Evaluated with \textbf{CD} metric ($10^{-2}).$ Lower is better.
    % Note that these baselines are reimplementations from VLN-CE~\cite{krantz2020navgraph} with small changes (see Section~\ref{baselines} for further details).\zk{Copied from prev. paper, needs update} 
    }
    
    \label{reconstruction_nocs}
    \resizebox{1.0\textwidth}{!}{
    \begin{tabular}{clcccccccccccccc}
        \toprule
        & & \multicolumn{7}{c}{\textbf{CAMERA25}} & \multicolumn{7}{c}{\textbf{REAL275}} \\ \cmidrule(r{0.1in}){3-9} \cmidrule(r{0.1in}){10-16}
        & {Method} & \textbf{Bottle} & \textbf{Bowl} & \textbf{Camera} & \textbf{Can} & \textbf{Laptop} & \textbf{Mug} & \textbf{Mean} & \textbf{Bottle} & \textbf{Bowl} & \textbf{Camera} & \textbf{Can} & \textbf{Laptop} & \textbf{Mug} & \textbf{Mean} \\
        \midrule
        1 & {Reconstruction~\cite{tian2020shape}} & 0.18 & 0.16  & 0.40 & 0.097 & 0.20 & 0.14 & 0.20& 0.34 &0.12 & 0.89 & 0.15 &0.29 &0.10 & 0.32 \\
        2 & {ShapePrior~\cite{tian2020shape}} &0.34 &0.22& 0.90& 0.22& 0.33& 0.21 &0.37 &0.50 &0.12& 0.99& 0.24& 0.71& 0.097& 0.44 \\
        3 & CenterSnap &0.11	&0.10&	0.29&	0.13&	0.07& 0.12	&	0.14 & 0.13 &  0.10 & 0.43 &0.09 & 0.07 & 0.06 & 0.15 \\
        \midrule
        3 & \textbf{\titleShort{}(Ours)} &0.14	&0.08&	0.2&	0.14&	0.07& 0.11	&	0.16 & 0.1 &  0.08 &0.4 &0.07 & 0.08 & 0.06 & 0.13 \\
        \bottomrule
    \end{tabular}
    }
    \vspace{-5mm}
\end{table*}

%% file: tables/optimization_3d.tex
\setlength{\tabcolsep}{6pt}
\begin{table}[t]
  \centering
  \small
  \renewcommand{\arraystretch}{1.3}
  \caption{\textbf{Generalizable Implicit Representation Ablation}: We evaluate the efficiency (point sampling/time(s)/memory(MB)) and generalization (shape(CD) and texture(PSNR) reconstruction) capabilities of our implicit object representation as well as its sampling efficiency for different levels of detail (LoDs) and compare it to the ordinary grid sampling. All ablations were executed on NVIDIA RTX A6000 GPU.}
  \resizebox{1.0\textwidth}{!}{%
    \begin{tabular}{c|c|cc|cc|cc}
    \toprule
    \multirow{2}[4]{*}{Grid type} & \multirow{2}[4]{*}{Resolution} & \multicolumn{2}{c|}{Point Sampling} & \multicolumn{2}{c|}{Efficiency (per object)} & \multicolumn{2}{c}{Reconstruction} \\
\cmidrule{3-8}          &       & Input & Output & Time (s) & Memory (MB) & Shape (CD) & Texture (PSNR) \\
    \midrule
    \multirow{3}[2]{*}{Ordinary} & 40    & 64000 & 412   & 10.96 & 3994  & 0.30  & 10.08 \\
          & 50    & 125000 & 835   & 18.78 & 5570  & 0.19  & 12.83 \\
          & 60    & 216000 & 1400  & 30.51 & 7850  & 0.33  & \textbf{19.52} \\
    \midrule
    \multirow{3}[2]{*}{OctGrid} & LoD5  & 1521  & 704   & \textbf{5.53} & \textbf{2376} & 0.19  & 9.27 \\
          & LoD6  & 5192  & 3228  & 6.88  & 2880  & \textbf{0.18} & 13.63 \\
          & LoD7  & 20246 & 13023 & 12.29 & 5848  & 0.24  & 16.14 \\
    \bottomrule
    \end{tabular}%
    }
  \label{tab:optimization_3d}
\end{table}%

%% file: sections/qualitative.tex
\section{Qualitative Results}
\begin{figure}[t]
   \centering
       \includegraphics[width=\linewidth]{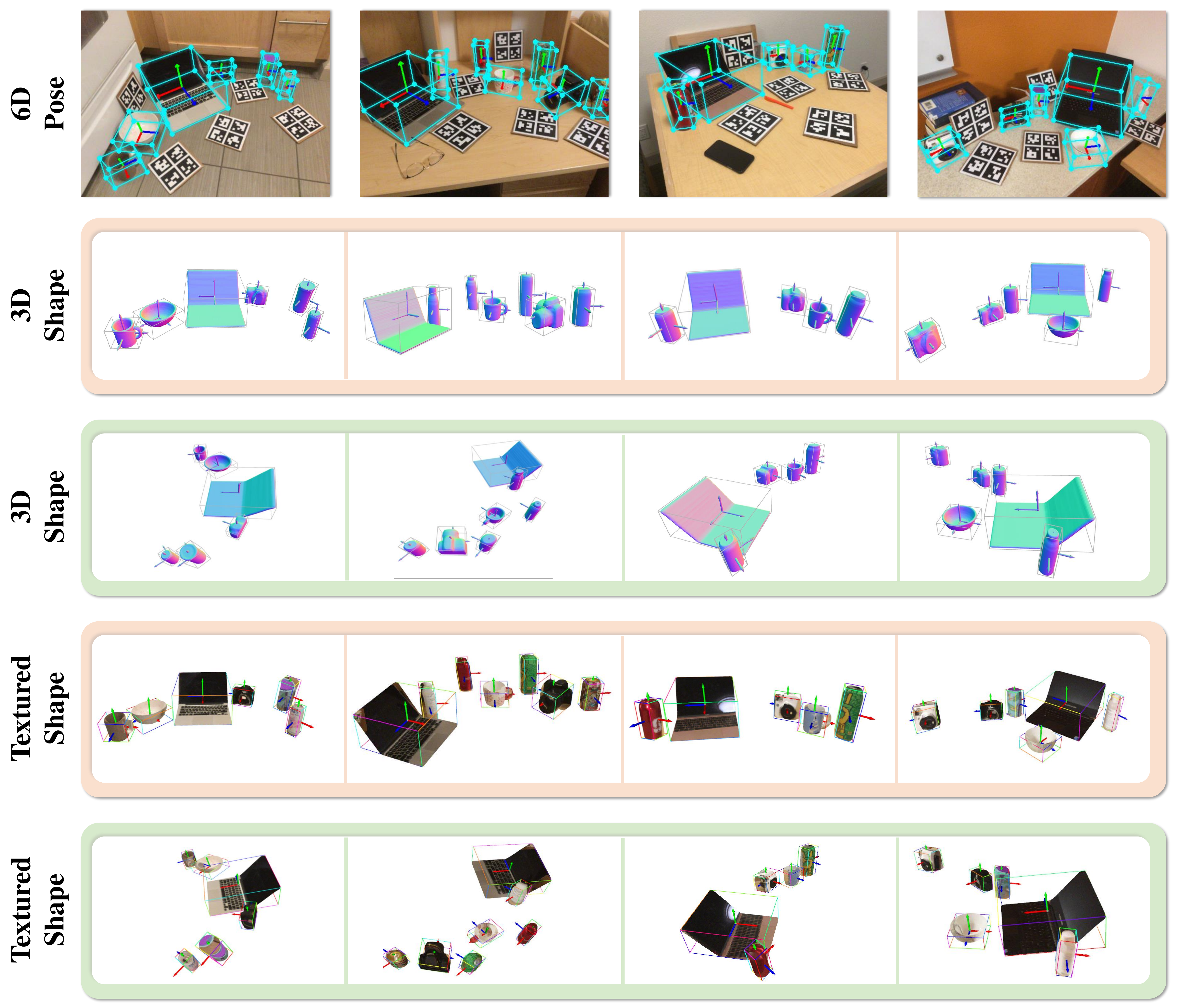}
      \caption{\textbf{\titleShort Qualitative 6D pose estimation and 3D Shape Reconstructions including Appearance:} Given a single-view RGB-D observation, our method reconstructs accurate 3D shapes along with appearances and estimates the 6D pose and sizes of multiple objects in the scene. Here, reconstructions are shown with different camera view-points i.e. orange and green backgrounds.} 
   \label{fig:qualitative_reconstructions}
\end{figure}

We qualitatively analyze the performance of our method \textbf{ShAPO} on the NOCS Real275 dataset~\cite{wang2019normalized}. As shown in Fig.~\ref{fig:qualitative_reconstructions}, our method reconstructs accurate 3D shapes and appearances of multiple novel objects along with estimating the 6D pose and sizes without requiring 3D CAD models of these novel instances (Results shown on 4 different real-world scenes containing novel
object instances using different camera-view points i.e. orange and green backgrounds). For further qualitative results, please consult supplementary materials. 

%% file: sections/conclusion.tex
\section{Conclusion}
In this paper we proposed ShAPO, an end-to-end method for joint multi-object detection, 3D textured reconstruction, 6D object pose and size estimation. Our method detects and reconstructs novel objects without having access to their ground truth 3D meshes. To facilitate this, we proposed a novel, generalizable shape and appearance space that yields accurate textured 3D reconstructions of objects in the wild. To alleviate sampling inefficiencies leading to increased time and memory requirements, we proposed a novel octree-based differentiable optimization procedure that is significantly more efficient than alternative grid based representations. For future work we will explore how the proposed method can be used to build object databases in new environments, to alleviate the cost and time required to construct high-quality 3D textured assets. A second avenue of future work consists of extensions to multi-view settings and integrating our work with SLAM pipelines for joint camera motion and object pose, shape and texture estimation in static and dynamic scenes.

%% file: sections/appendix.tex
\setcounter{section}{0}
\setcounter{figure}{0}
\setcounter{table}{0}

\looseness=-1
\section*{Appendix A: Qualitative Results Visualization}
\begin{figure}[htbp]
   \centering
       \includegraphics[width=\linewidth]{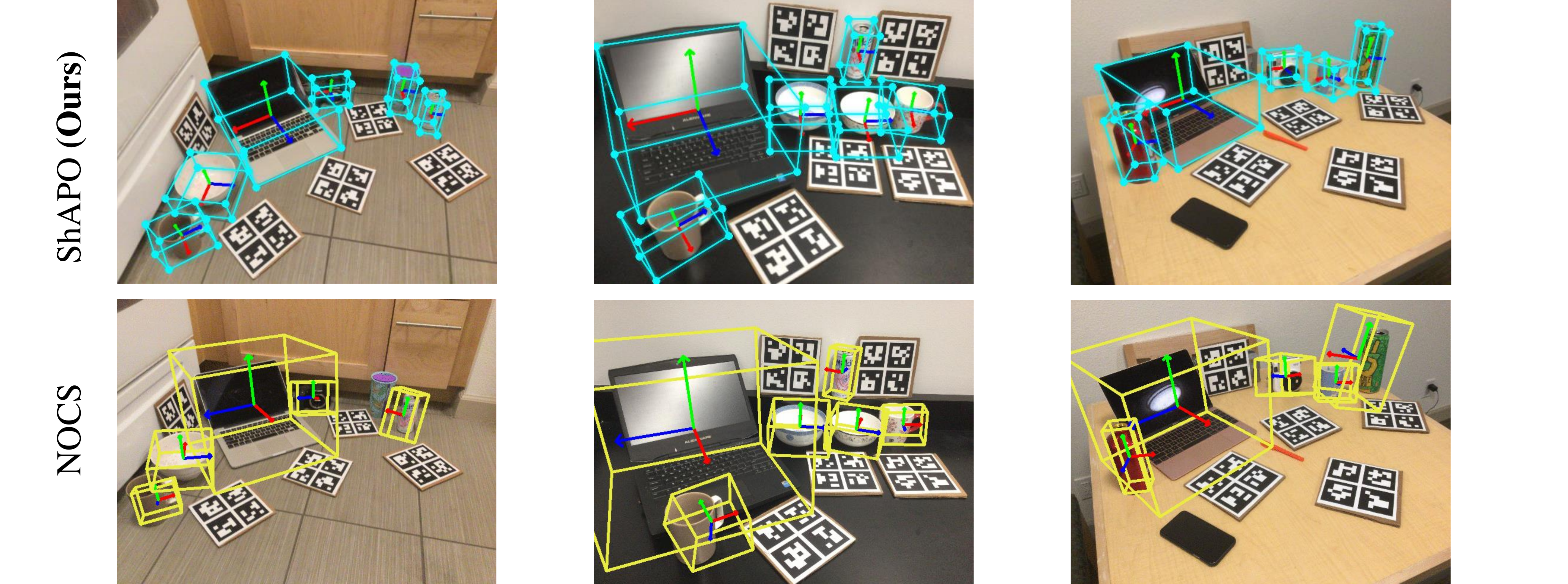}
      \caption{\textbf{\titleShort Qualitative 6D pose estimation comparison with NOCS:} Our method's 6D pose estimation in comparison to the best pose estimation configuration i.e. 32-bin classification on the NOCS dataset. We show accurate 3D bounding box predictions and 6D pose and size estimation of multiple novel object categories than the strong baseline.} 
   \label{fig:nocs_comparison}
\end{figure}

\begin{figure}[htbp]
   \centering
       \includegraphics[width=\linewidth]{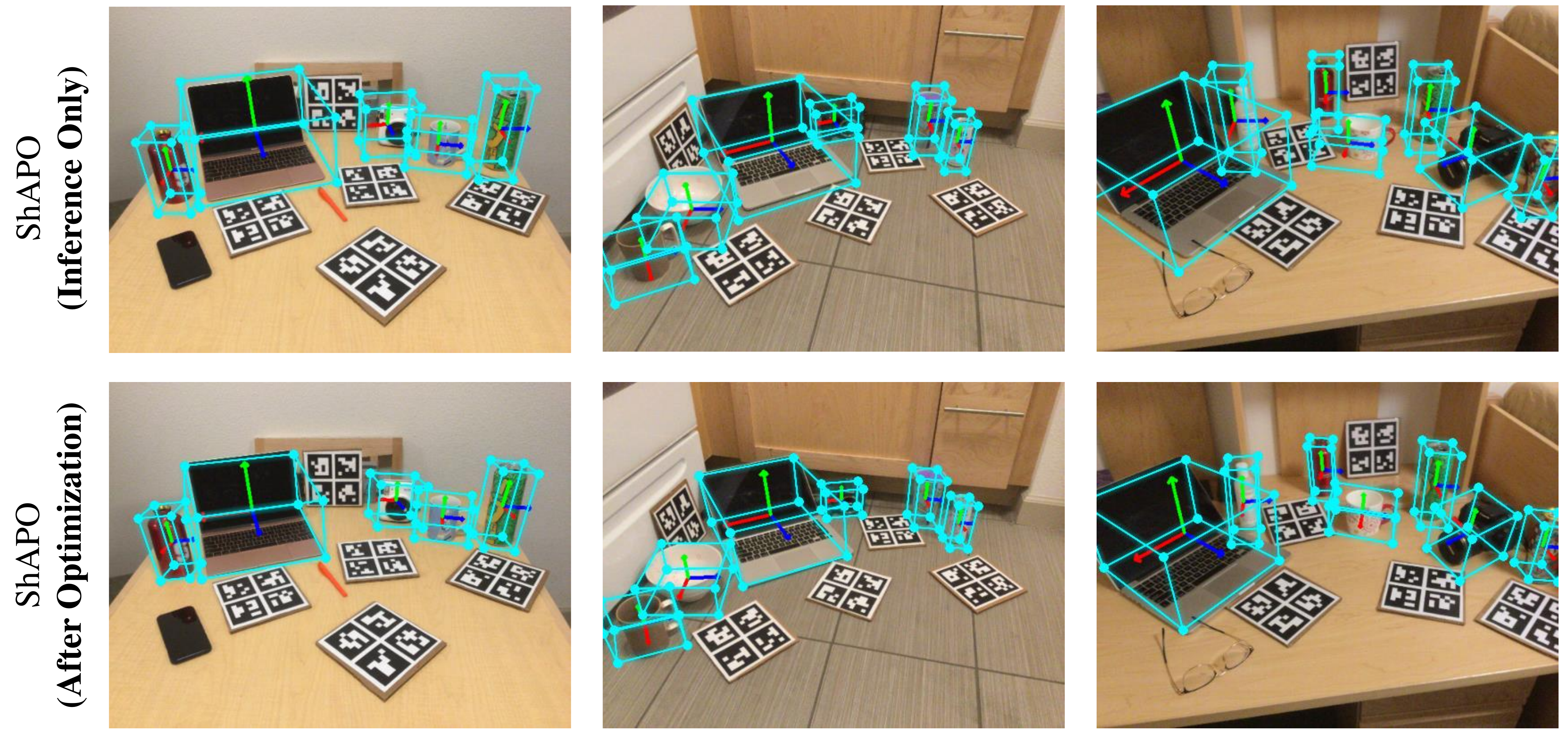}
      \caption{\textbf{\titleShort Qualitative Comparison of 6D pose and size Inference and Optimization:} Our method's 6D pose and size comparison shown on 3 novel scenes in NOCS Real275 test-set. After optimization, our method predicts accurate bounding boxes as shown by the bottom row in the figure.} 
   \label{fig:before_after_optim_pose}
\end{figure}

\begin{figure}[htbp]
   \centering
       \includegraphics[width=\linewidth]{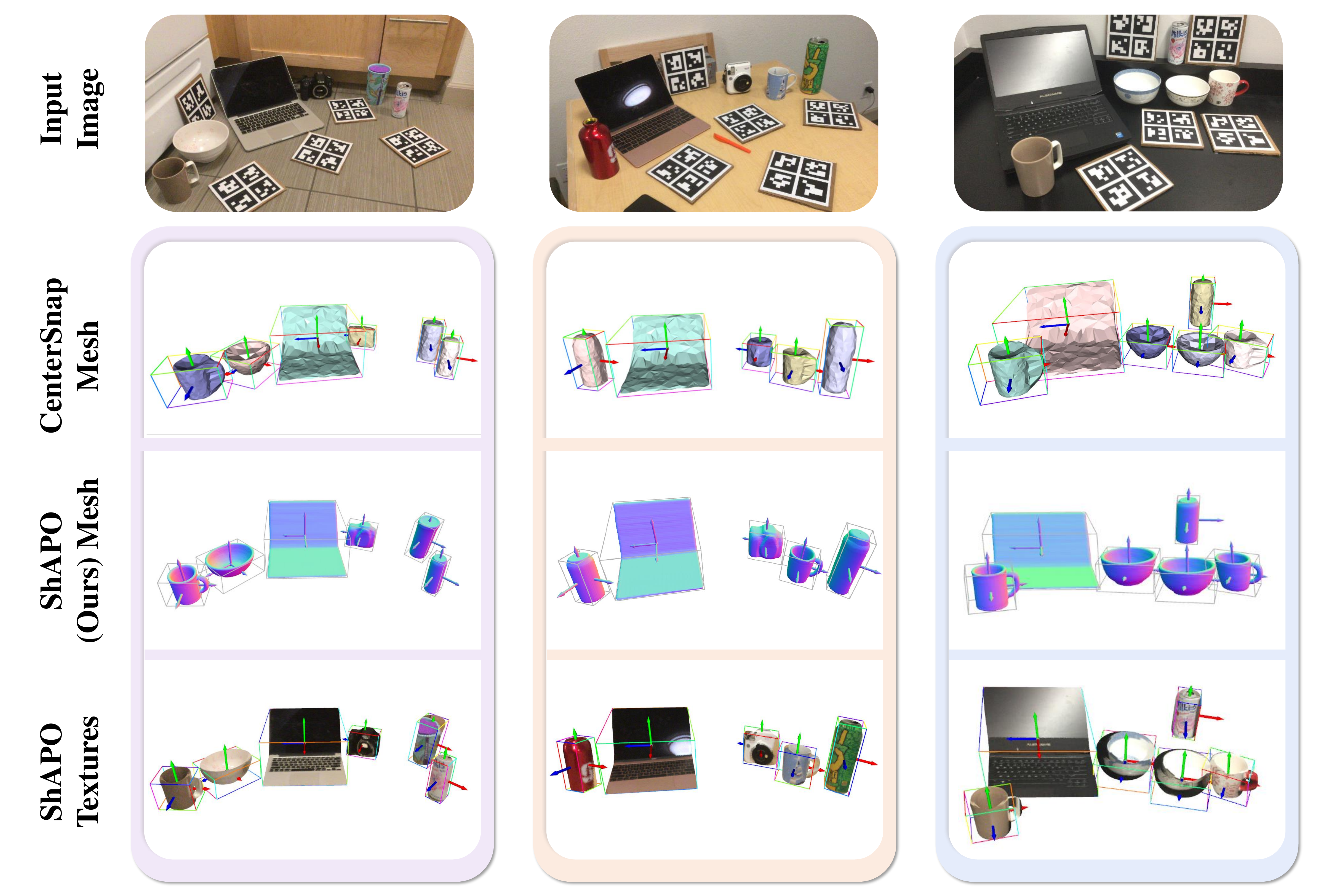}
      \caption{\textbf{\titleShort Qualitative Reconstruction Comparison with CenterSnap~\cite{irshad2022centersnap}:} The figure qualitatively shows the superior reconstruction performance of our method with the strong state of art i.e. CenterSnap~\cite{irshad2022centersnap} on novel scene in NOCS Real275 test-set. Our method produces finer reconstruction surfaces both in terms of shape accuracy and textures with details such as mug-handle and camera lens.} 
   \label{fig:centersnap_comparison}
\end{figure}

\begin{figure}[htbp]
\begin{center}
\includegraphics[width=0.92\linewidth]{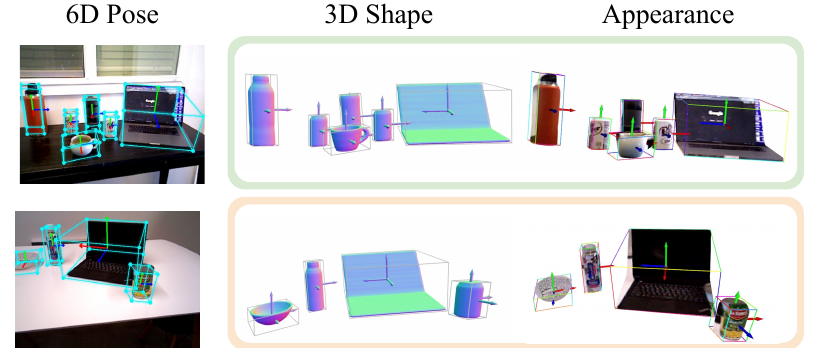}
\captionof{figure}{\textbf{Zero-shot real-world generalization experiments on HSR robot}}
  \label{fig:HSR}
\end{center}
\end{figure}

Here we provide more visual qualitative result of superior single-view multi-object \textit{Shape Reconstruction}, \textit{6D pose and size estimation} and \textit{Appearance Reconstruction} done using our technique, \textbf{ShaPO}. Our method shows very promising results for superior 6D pose and size estimation compared to the strong baseline NOCS~\cite{wang2019normalized}~(Figure~\ref{fig:nocs_comparison}). Our network also performs more accurate shape and texture reconstruction compared to the strong-baseline, CenterSnap~\cite{irshad2022centersnap}~(Figure ~\ref{fig:centersnap_comparison}), which only performs shape reconstruction~(i.e. meshes obtained through surface reconstruction of coarse pointcloud predictions i.e. 2048 points). We also visualize the improved pose estimation performance of our method after inference-time optimization~(Figure~\ref{fig:before_after_optim_pose}). Figure~\ref{fig:HSR} also shows zero-shot generalization results on HSR robot i.e. no re-training was done. 

\vspace{-1cm}
\setlength{\tabcolsep}{16pt}
\begin{table}[htbp]
  \centering
  \caption{\textbf{Texture quality ablation.} We compare texture quality using the PSNR metric between three modalities: network prediction, optimization, and fine-tuning of the $t_{\theta}$ network.}
    \resizebox{0.8\textwidth}{!}{%
    \begin{tabular}{l|ccc}
    \toprule
          & \multicolumn{1}{l}{Inference} & \multicolumn{1}{l}{Optimization} & \multicolumn{1}{l}{Fine-tuning} \\
    \midrule
    PSNR  & 11.41 & 20.64 & 24.32 \\
    \bottomrule
    \end{tabular}%
    }
  \label{tab:psnr}%
\end{table}%

\looseness=-1
\section*{Appendix B: Texture Quality Ablation on NOCS Real275}
\begin{figure}[htbp]
   \centering
       \includegraphics[width=\linewidth]{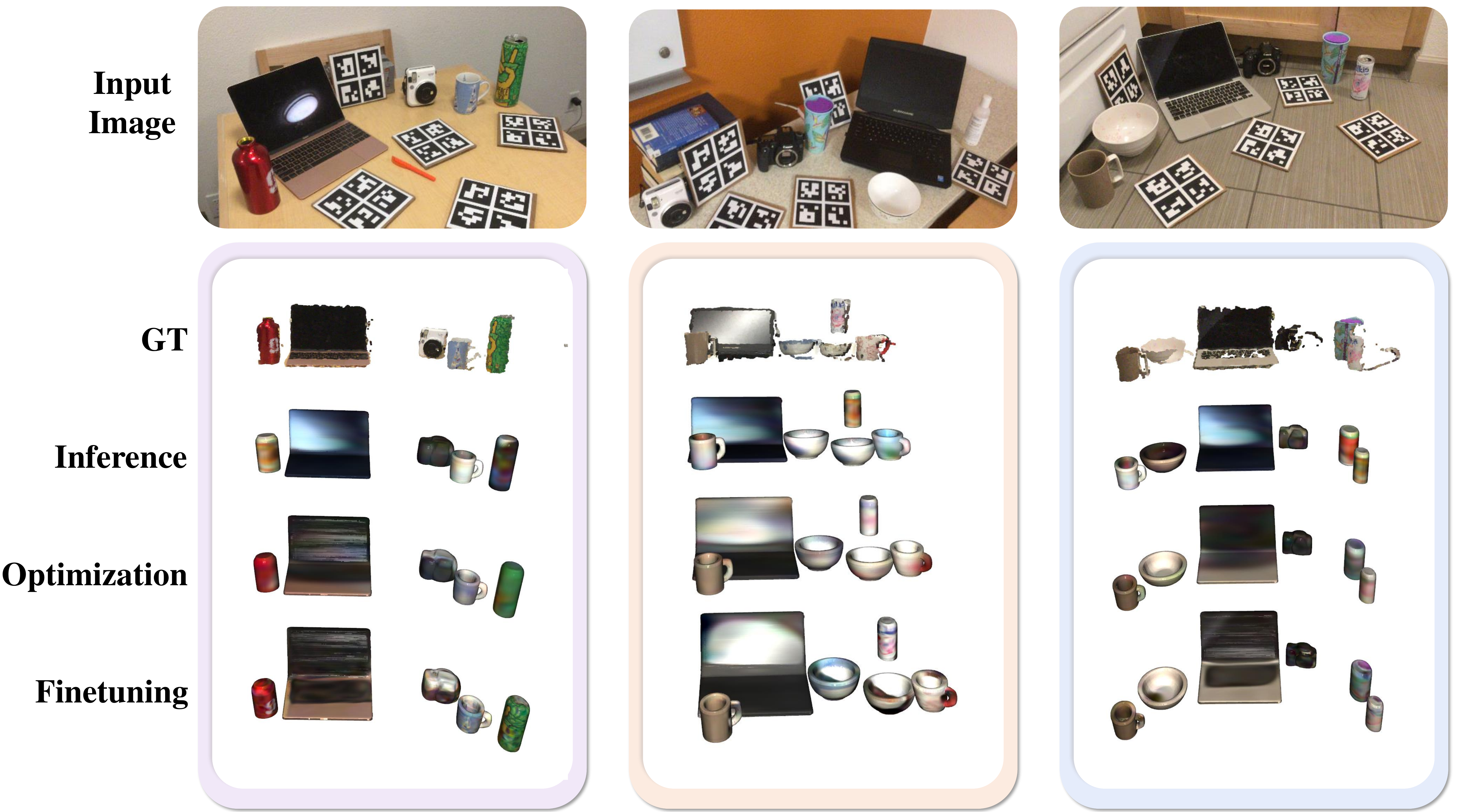}
      \caption{\textbf{\titleShort Qualitative Inference, Optimization and Finetuning Comparison:} The figure qualitatively shows the inference, latent-only optimization and latent with appearance network optimization. Note that as noted earlier, we let the appearance network weights to change to allow for finer level of reconstruction. } 
   \label{fig:inference_optim_comparison}
\end{figure}

In this section, we provide an ablation on the output texture quality on NOCS Real275 test-scenes. In particular, we compare the direct network texture prediction with the result after our differentiable optimization, and the result after our differentiable optimization with additional fine-tuning of the $t_{\theta}$ network weights. We use the learning rate of $10^{-5}$ for the weight fine-tuning. Table~\ref{tab:psnr} demonstrates that our optimization procedure almost doubles the texture quality in terms of PSNR. Additional fine-tuning of the network weights allows us to improve texture reconstruction results even further. For qualitative results see Figure~\ref{fig:inference_optim_comparison}.


\section*{Appendix C: Network Architecture details and Training}
Our backbone is implemented as Feature pyramid network~\cite{kirillov2019panoptic} with takes as input Resnet~\cite{heDeepResidualLearning2016} outputs at various spatial resolutions and adds lateral connections with a top-down pathway. Each of our specialized heads comprises of a series of convolution layers (i.e. up-sampling stages), as described in~\cite{Kirillov_2019_CVPR} with the final prediction layer comprising of a $1\times1$ convolution and $4x$ bilinear up-sampling. We train the combined backbone and heads network for 30 epochs with early stopping based on the performance on the validation set. We use a learning rate of $6e^{-4}$ and a polynomial weight decay with a co-efficient of $1e^{-4}$. Our texture network $t_{\theta}$ is a Siren-based~\cite{sitzmann2020implicit} 6-layer MLP consisting of 512-dimensional hidden layers and with $\omega_0$ set to 128. Siren networks demonstrate superior results at representing fine details when compared to standard ReLU-based MLPs thanks to the used periodic activation functions. After we train the shape MLP~($G$) and texture MLP~($t_{\theta}$), we freeze the networks for single-shot supervision at the Gaussian center locations. During inference, we use the frozen networks~(G) and~($t_{\theta}$) to optimize for shape, pose, size and appearance latent codes.

\section*{Appendix D: Related works} 
Our method, ShAPO, relates to multiple key areas in 3D scene reconstruction, object understanding and pose estimation. In essence, we provide more qualitative and quantitative comparison to four related works i.e. CenterSnap~\cite{irshad2022centersnap}, MeshSDF~\cite{remelli2020meshsdf}, Occ-Nets~\cite{mescheder2019occupancy} and TextureFieldd~\cite{OechsleICCV2019}. \\
~\textbf{CenterSnap~\cite{irshad2022centersnap}}: In particular, Figure~\ref{fig:centersnap_comparison} qualitatively shows the superior reconstruction quality of our method compared to the strong state of the art i.e. CenterSnap~\cite{irshad2022centersnap}. Furthermore, our shape representation (implicit SDF vs pointclouds in CS), addition of texture network and texture codes, differentiable iso-surface extraction, optimization and joint shape, appearance, and textures warping all make our work significantly different from CS. While our work does share a common backbone with CS, being able to leverage the test-time observations to warp the latent shape, appearance, and poses of the model beyond network inference is precisely our contribution. Hence, our technique is able to model large intra-class variations (25.4\% and 7.1\% absolute improvement in 6D pose) over CS (cf. Tbl 2 in the main text). \\
~\textbf{MeshSDF~\cite{remelli2020meshsdf}} proposes a solution to extracting surface meshes while preserving end-to-end differentiability. We instead extract dense surface point clouds using our octree-based sampling abstaining from the expensive Marching Cubes computation at every optimization step. As shown in the Table~\ref{tab:meshsdf}, MeshSDF’s implementation doesn’t scale well to higher resolutions, whereas our technique does, achieving accurate fine-grained reconstruction with minimal runtime. Additionally, our method also supports appearance optimization. \\
~\textbf{Occupancy Networks~\cite{mescheder2019occupancy}}: We extract the object's surface using a \textbf{differentiable 0-isosurface projection} which is a crucial component that allows us to perform shape/pose/appearance optimization. Conversely, the procedure from Occupancy Networks~\cite{mescheder2019occupancy} is applied once to extract the object's surface using non-differentiable Marching Cubes. \\
~\textbf{Texture Fields~(TF)~\cite{OechsleICCV2019}} trains a single network per category making it difficult to model a large number of categories. We model multiple categories through our novel texture code ($\textbf{z}_{tex}$) unique to each object in our database of shape and texture priors using a single network (cf. supplementary video). Second, TF does not consider test-time optimization, whereas we propose a novel test-time warping of textures by updating $\textbf{z}_{tex}$ to fit unseen object appearances. Lastly, TF reconstructs one model per image whereas we infer multiple objects from a single-view RGBD.

\setlength{\tabcolsep}{14pt}
\begin{table}[t]
\vspace{-15pt}
  \centering
%   \caption{Add caption}
  \caption{\textbf{Quantitative Comparison with MeshSDF} We compare the computational time to extract surface for our method, ShAPO, in comaprison to MeshSDF~\cite{remelli2020meshsdf}}
    \resizebox{1\columnwidth}{!}{
    \begin{tabular}{c|ccccc}
    \toprule
    \# Points & 704   & 3228  & 13023 & 56041 & 224680 \\
    \midrule
    \textbf{MeshSDF} & 0.017 s & 0.032 s & 0.091 s & 0.654 s & 4.396 s \\
    \textbf{ShaPO} & 0.010 s & 0.013 s & 0.016 s & 0.025 s & 0.093 s \\
    \bottomrule
    \end{tabular}%
    }
  \label{tab:meshsdf}%
\end{table}%

\section*{Appendix E: Failure Modes, Limitations and Future Work}
\begin{figure}[htbp]
\begin{center}
\includegraphics[width=0.75\linewidth]{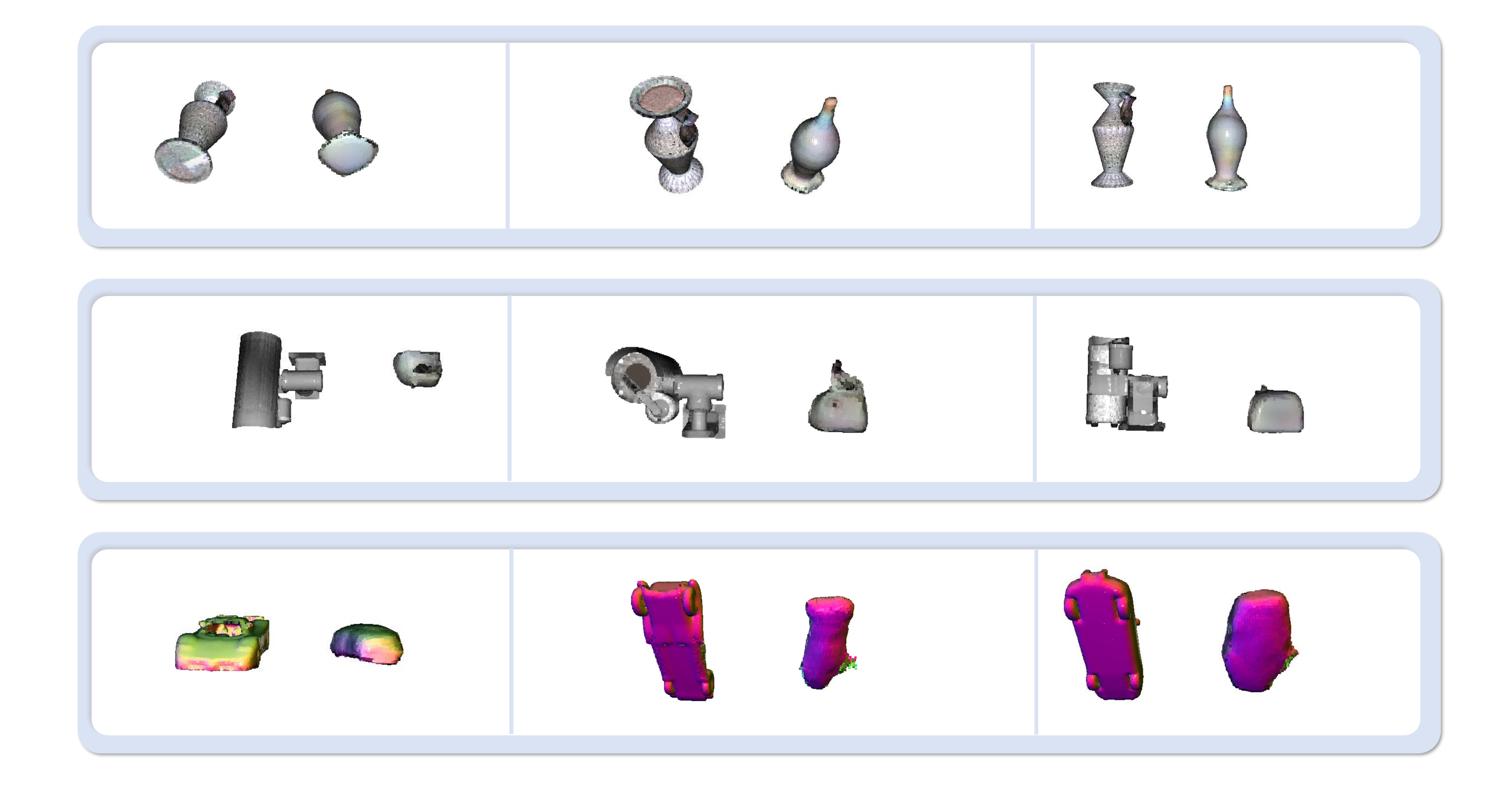}
\captionof{figure}{\textbf{Out-of-distribution failure modes}}
  \label{fig:failure_modes}
\end{center}
\end{figure}

We further qualitatively analyze the performance of our 3D shape optimization on out-of-distribution scenarios, on three categories not present in the training set, i.e. vase, car and engine. As shown in Figure~\ref{fig:failure_modes}, we show that since our shape MLP has not seen these categories during training, it is difficult for the model to optimize towards a new observation based on a single optimization alone. Hence, a strong prior is needed over a large database of shapes to generalize to a new instance from a single observation. This further confirms the methodology of ShAPO where we build a strong prior using a differentiable implicit shape and textured database. Note that generalization to new categories~(i.e. not seen during training) is possible using our method and
it requires adding 3D CAD models during the shape and texture pre-training stage. We believe that scaling our method ShAPO to a large number of categories is an interesting direction and we leave this to future work. We also believe another interesting direction is to use this method, specially the differentiable optimization, for self-supervised 6D pose and size estimation in-the-wild, i.e. not requiring any 6D pose and size labels for real-world scenarios. 
